\documentclass[sigconf]{acmart}

\AtBeginDocument{%
  \providecommand\BibTeX{{%
    \normalfont B\kern-0.5em{\scshape i\kern-0.25em b}\kern-0.8em\TeX}}}

\setcopyright{acmcopyright}
\copyrightyear{2018}
\acmYear{2018}
\acmDOI{XXXXXXX.XXXXXXX}

\acmConference[Conference acronym 'XX]{Make sure to enter the correct
  conference title from your rights confirmation emai}{June 03--05,
  2018}{Woodstock, NY}
\acmPrice{15.00}
\acmISBN{978-1-4503-XXXX-X/18/06}

\usepackage{multirow}
\usepackage{url}
\usepackage{xcolor}

\begin{document}

\title{Mining Knowledge from Query Image for Few Shot Remote Sensing Image Semantic Segmentation}
\title{Self-Correlation and Cross-Correlation Learning for Few-Shot Remote Sensing Image Semantic Segmentation}

\author{Linhan Wang}
\email{linhan@vt.edu}
\affiliation{%
  \institution{Virginia Tech}
  \state{Virginia}
  \country{United States}
}

\author{Shuo Lei}
\email{slei@vt.edu}
\affiliation{%
  \institution{Virginia Tech}
  \state{Virginia}
  \country{United States}
}

\author{Jianfeng He}
\email{jianfenghe@vt.edu}
\affiliation{%
  \institution{Virginia Tech}
  \state{Virginia}
  \country{United States}
}

\author{Shengkun Wang}
\email{shengkun@vt.edu}
\affiliation{%
  \institution{Virginia Tech}
  \state{Virginia}
  \country{United States}
}

\author{Min Zhang}
\email{minzhang23@vt.edu}
\affiliation{%
  \institution{Virginia Tech}
  \state{Virginia}
  \country{United States}
}

\author{Chang-Tien Lu}
  \email{ctlu@vt.edu}
\affiliation{%
  \institution{Virginia Tech}
  \state{Virginia}
  \country{United States}}

\renewcommand{\shortauthors}{Wang, et al.}

\begin{abstract}
Remote sensing image semantic segmentation is an important problem for remote sensing image interpretation. Although remarkable progress has been achieved, existing deep neural network methods suffer from the reliance on massive training data. Few-shot remote sensing semantic segmentation aims at learning to segment target objects from a query image using only a few annotated support images of the target class. Most existing few-shot learning methods stem primarily from their sole focus on extracting information from support images, thereby failing to effectively address the large variance in appearance and scales of geographic objects. To tackle these challenges, we propose a \textbf{S}elf-\textbf{C}orrelation and \textbf{C}ross-Correlation Learning \textbf{Net}work for the few-shot remote sensing image semantic segmentation. Our model enhances the generalization by considering both self-correlation and cross-correlation between support and query images to make segmentation predictions. To further explore the self-correlation with the query image, we propose to adopt a classical spectral method to produce a class-agnostic segmentation mask based on the basic visual information of the image. Extensive experiments on two remote sensing image datasets demonstrate the effectiveness and superiority of our model in few-shot remote sensing image semantic segmentation. Code and models will be accessed at \url{https://github.com/linhanwang/SCCNet}.
\end{abstract}



\received{20 February 2007}
\received[revised]{12 March 2009}
\received[accepted]{5 June 2009}

\maketitle

\section{Introduction}
Semantic segmentation in remote sensing images has become an essential task for various applications, such as land use analysis \cite{ienco2019combining}, urban management \cite{wurm2019semantic}, environmental monitoring \cite{kattenborn2021review}, and other areas of national economic development. 
Although deep neural networks for semantic segmentation~\cite{long2015fully,badrinarayanan2017segnet,chen2017deeplab,zhao2017pyramid} have achieved remarkable progress, their reliance on large-scale datasets greatly restricts their application in low-resource domains. For example, collecting an adequate amount of remote sensing data is hard, and the expense associated with hiring domain experts to annotate the data is too costly to be feasible.
To reduce such burden on data annotation, few-shot semantic segmentation has been proposed~\cite{shaban2017one}, which aims to learn a model that can perform segmentation on novel classes with only a few annotated images.

\begin{figure}[htb]
  \centering
  \includegraphics[width=\columnwidth]{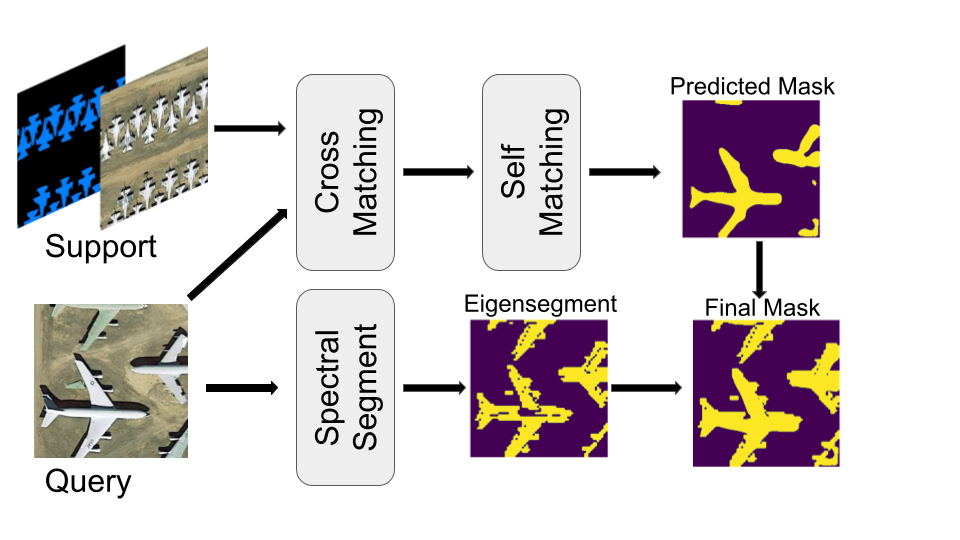}
  \caption{Overview of our proposed model (SCCNet) for few-shot remote sensing image semantic segmentation. SCCNet incorporates self-correlation information into the model and decomposes an image using the eigenvectors of a Laplacian matrix constructed from a visual feature to extract self-contained knowledge from the query image.}
  \label{fig:motivation}
\end{figure}

Recently, a group of few-shot segmentation methods adopted global average pooling \cite{shaban2017one} over the foreground region of the support features to generate class prototypes, which are then employed to guide the segmentation process of the query image. Building upon this research direction, some studies \cite{wang2019panet, yang2020prototype} strive to design more representative support prototypes to enhance segmentation performance. While significant advancements have been made for natural images, these methods encounter challenges when applied to remote sensing images, primarily due to the presence of large intra-class variances. Specifically, geographic objects of the same class can exhibit substantial variations in appearance and scales from different angles. Recently, SDM \cite{yao2021scale} proposes scale-aware focal loss to focus training on tiny hard-parsed objects and performs detailed matching with multiple prototypes for providing more accurate parsing guidance. However, SDM only considers the cross-correlation between support and query images, ignoring the self-correlation between pixels within the query image. We found that self-correlations within the query image could provide extra knowledge to help segment tiny objects, which is very important for few-shot remote sensing image semantic segmentation, particularly when there is a significant discrepancy between the support and query images. 

To address the aforementioned challenges, we propose a novel model, named \emph{SCCNet}, to leverage knowledge from query images for few-shot remote sensing image semantic segmentation. As illustrated in Fig. \ref{fig:motivation}, the proposed model consists of two key components. First, we incorporate the initial query mask prediction to collect query features in high-confidence regions and then use the generated masked query features to perform self-matching with query
features. Since pixels belonging to the same object are expected to exhibit higher similarity than those belonging to different objects,
Self-Matching Module can provide auxiliary support information to segment the query image.
Second, we propose a novel Spectral Segmentation Module to extract knowledge from query images further with classical spectral methods. Specifically, we first construct the affinity matrices using basic visual information (i.e. color and position information) and semantic information derived from the middle-layer features of the pretrained backbone.
Then we decompose images using the eigenvectors of Laplacian of affinity matrices as soft segments and obtain the class-agnostic eigensegments. Since it operates solely on the query images without relying on the support annotations, it is naturally resilient to the significant discrepancies that may exist between the support and query images. The final prediction mask of the query image is obtained by fusing the optimized query mask and the eigensegment.

Our key contributions can be summarized as follows:
\begin{itemize}
    \item We propose a \textbf{S}elf-\textbf{C}orrelation and \textbf{C}ross-Correlation Learning \textbf{Net}work for the few-shot remote sensing image semantic segmentation. Our model enhances the generalization by considering both self-correlation within query images and cross-correlation between support and query images to make segmentation predictions.
    \item We proposed a Self-Matching Module to extract more comprehensive query information. The correlation between the initial segment and the query images is introduced to the model to tackle the large discrepancy between support and query images.
    \item We propose a novel Spectral Segmentation Module with spectral analysis to produce class-agnostic segmentations of query images without the supervision of any annotations.
    \item We evaluate the proposed model on two remote sensing image datasets for few-shot semantic segmentation tasks. Comprehensive experiments demonstrate that our \emph{SCCNet} consistently outperforms all the baselines for both 1-shot and 5-shot settings.
\end{itemize}

\section{Related Work}

\subsection{Remote Sensing Image Semantic Segmentation}

Deep learning-based methods have gained significant popularity in the remote sensing community, showcasing remarkable progress in segmenting remote sensing images. Specifically, Maggiori et al. \cite{maggiori2017high} introduced a multilayer perceptron (MLP) into the segmentation network to produce better segmentation
results. Yu et al. \cite{yu2018semantic} introduced the pyramid pooling module as a means to address semantic segmentation in remote sensing images, while Yue et al. \cite{yue2019treeunet} developed TreeUNet as the first adaptive Convolutional Neural Network (CNN) specifically tailored for semantic segmentation in this domain. Zhang et al. \cite{zhang2020multi} adopted the multibranch parallel convolution structure in HRNet \cite{sun2019high} to generate multiscale feature maps and designed an adaptive spatial pooling module to aggregate more local contexts. To tackle the challenge in small-scale object segmentation, Kammpffmeyer et al. \cite{kampffmeyer2016semantic} assembled patch-based pixel classification and pixel-to-pixel segmentation, which introduced uncertain mapping to achieve high performance on small-scale objects. FactSeg \cite{ma2021factseg} proposed a symmetrical dual-branch decoder consisting of a foreground activation branch and a semantic refinement branch. The two branches performed multiscale feature fusion through skip connection, thereby improving the accuracy of small-scale object segmentation. Furthermore, with the emergence of multiple attention mechanisms, Ding et al. \cite{ding2020lanet} designed an efficient local attention embedding to enhance segmentation performance. 

Although existing methods effectively demonstrate the capabilities of deep learning in remote sensing image semantic segmentation, they typically require a large number of densely-annotated images for training and have difficulties in generalizing
to unseen object categories.

\subsection{Few-shot Semantic Segmentation} 
To address the generalization issue and reduce massive training data annotation, Few-Shot Semantic Segmentation (FSS) task has been proposed, which aims to learn a model that can perform segmentation on novel classes with only a few pixel-level annotated images. Shaban et al. \cite{shaban2017one} first proposed one-shot semantic segmentation networks to address FSS. It uses global average pooling over the foreground region of the support
features to generate class prototypes, which are then employed
to guide the segmentation process of the query image. Building upon the concept of prototypical networks~\cite{snell2017prototypical}, utilizing prototype representations to guide mask prediction in query images has become a popular paradigm in the field of few-shot segmentation. Specifically, PANet \cite{wang2019panet} proposed a prototype alignment regularization between support and query images to generate high-quality prototypes. PMMs \cite{yang2020prototype} employ the Expectation-Maximization algorithm to generate multiple prototypes corresponding to different parts of the objects. 
Recently, a group of matching-based methods has been proposed to leverage dense correspondences between query images and support annotations. HSNet \cite{min2021hypercorrelation} utilizes 4D convolutions to extract precise segmentation masks by compressing the multilevel 4D correlation tensors. VAT \cite{hong2022cost} proposes a 4D Convolutional Swin Transformer to aggregate the correlation map. To fully harness the information within the support set, Yang et al. \cite{yang2021mining} employ clustering techniques to mine latent novel classes in the support set and subsequently treat them as pseudo labels during the training process.

Despite the remarkable progress achieved in natural images, Yao et al. \cite{yao2021scale} found that performance drops
dramatically on unseen classes in remote sensing images. This limitation arises from the inability of these methods to effectively handle the significant variations in object appearance and scales prevalent in remote sensing images. To address this challenge, SDM \cite{yao2021scale} proposes a scaled-aware focal loss, which enhances the focus on tiny objects. DMML-Net \cite{wang2021dmml} uses an affinity-based fusion mechanism to adaptively calibrate the deviation of the prototype induced by intra-class variation.

It is worth noting that all existing methods primarily focus on extracting information solely from the support set to make a segmentation. However, we argue that this approach may not be sufficient for remote sensing images, where substantial discrepancies exist between the support and query images. In this study, we aim to pioneer a novel direction by extracting the self-contained knowledge in the query images to boost the performance for few-shot remote sensing image semantic segmentation.

\subsection{Spectral Methods for Segmentation}

Spectral analysis originally emerged from the exploration of continuous operators on Riemannian manifolds \cite{cheeger1970lower}. Subsequent research efforts extended this line of research to the discrete setting of graphs, leading to numerous findings that connect global graph properties to the eigenvalues and eigenvectors of their associated Laplacian matrices. Lin et al. \cite{lin2020space} demonstrate that the eigenvectors of graph Laplacians yield graph partitions with minimum energy. Building upon this insight, Shi et al. \cite{shi2000normalized} view image segmentation as a graph partitioning problem and propose a novel global criterion called the normalized cut for image segmentation. As presented by Aksoy et al. \cite{aksoy2018semantic}, soft segmentations are automatically generated by fusing high-level and low-level image features within a graph structure. The construction of this graph facilitates the utilization of the corresponding Laplacian matrix and its eigenvectors to reveal semantic objects and capture soft transitions between them.


\section{Problem Setup}

Few-shot semantic segmentation aims to perform segmentation on the novel classes with only a few annotated images. Suppose we are provided with images from two non-overlapping class sets: $\mathcal{C}_{base}$ and $\mathcal{C}_{novel}$.
The training dataset $\mathcal{D}_{train}$ is constructed from the class set $\mathcal{C}_{base}$ and the test dataset $\mathcal{D}_{test}$ is constructed from the class set $\mathcal{C}_{novel}$.

To mitigate the risk of overfitting caused by limited training data, we adopt a commonly used meta-learning technique known as episodic training \cite{vinyals2016matching}. In the $K$-shot setting, we employ episodic sampling to select $K+1$ annotated image pairs, denoted as $\{(I_1^s, M_1^s), \newline (I_2^s, M_2^s), ..., (I_K^s, M_K^s), (I^q, M^q)\}$, with the same targeted class from the training dataset $\mathcal{D}_{train}$. Here, $\{(I_i^s, M_i^s)\}_{i=1}^K$ represents the support samples, and $(I^q, M^q)$ denotes the query pair. During the training phase, the segmentation model takes both the support samples $\{(I_i^s, M_i^s)\}_{i=1}^K$ and the query image $I^q$ as inputs and generates a predicted mask $\tilde{M}^q$. This prediction is then supervised by the corresponding ground truth mask $M^q$. Similarly, during the testing phase, we employ $K$ annotated image pairs from $\mathcal{D}_{test}$ to infer the semantic objects present in the query images. 

\section{Proposed Approach}
\begin{figure*}[tb]
  \centering
  \scalebox{0.8}{
  \includegraphics[trim={0.5cm 1.5cm 0cm 2.5cm},clip]{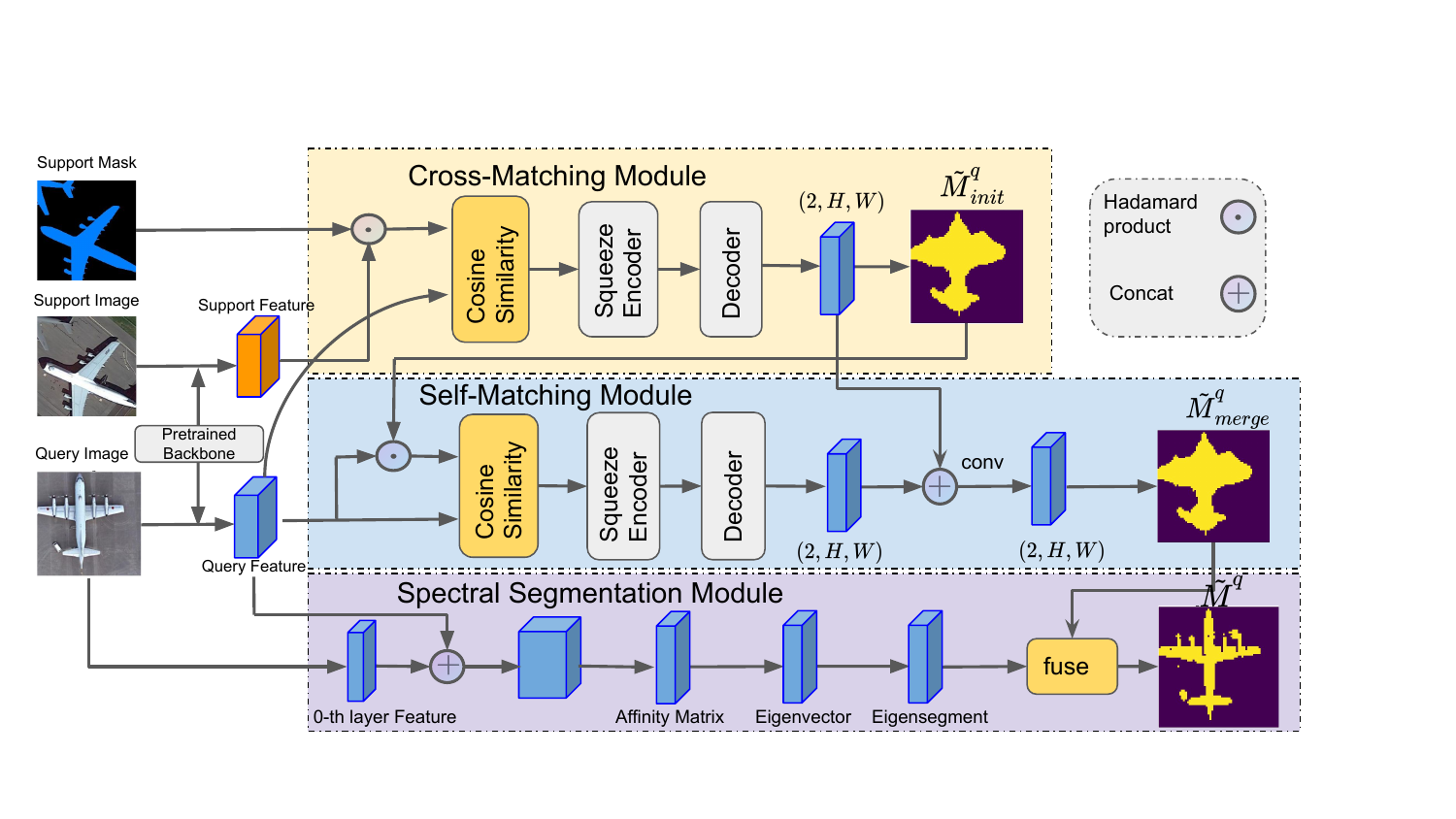}}
  \caption{Overall pipeline of our proposed network. The pretrained backbone is utilized as the feature extractor to generate corresponding support and query features. In the Cross-Matching Module, correlation between masked support features and query features is built and squeezed to generate the initial query mask. In the Self-Matching Module, correlation between query image and the initial query mask is further added into the model to generate finer query mask. Finally in the Spectral Segmentation Module, the query mask is fused with eigensegments obtained by non-learning-based spectral analysis.}
  \Description{Overall model structure}
  \label{fig:structure}
\end{figure*}

To solve the few-shot semantic segmentation problem in remote sensing images, we propose a novel model named \emph{SCCNet} as shown in Fig \ref{fig:structure}. First, we use pre-trained CNNs (VGG \cite{simonyan2014very} or Resnet \cite{he2016deep}) as the feature extractor to generate the corresponding query and support features. In the cross-matching module, pixel-wise multi-scale correlation tensors between masked support features and query features are built and squeezed to generate the initial predicted query mask $\tilde{M}_{init}^q$. 
To tackle the high intra-class variance problem in remote sensing images, the Self-Matching Module calculates the correlations between query features masked by $\tilde{M}_{init}^q$ and other query features. These correlations are further squeezed and merged with squeezed correlations between support and query features to generate optimized query mask $\tilde{M}_{merge}^q$. To further mine knowledge from the query images, in the spectral segmentation module, the classic spectral analysis method is utilized to exploit the proximity of local regions. Specifically, eigenvectors of the Laplacian of the affinity matrix are utilized as soft segments and transformed into eigensegments by Thresholding algorithms afterward. In the end, the final prediction mask of the query image is obtained by fusing the optimized query mask and the eigensegment.

\subsection{Cross-Matching Module}
Different from encoding an annotated support image to a feature vector to facilitate query image segmentation, we adopt the pixel-wise correlation between the support and query images to make a segmentation in our Cross-Matching Module.

\noindent \textbf{Hypercorrelation pyramid construction.} We extract features from query and support images and compute the correlation between them. Given a pair of query and support images, $I^q$ and $I^s$, we adopt a pretrained backbone to produce a sequence of $L$ feature maps, $\{(F_l^q, F_l^s)\}_{l=1}^L$, where $F_l^q$ and $F_l^s$ denote query and support feature maps at the $l$-th level, respectively. A support mask $M^q$ is used to encode segmentation information and filter out the background information. We obtain a masked support feature as $\hat{F}_l^s = F_l^s \odot \zeta_l(M^s)$, where $\odot$ denotes the Hadamard product and $\zeta_l : \mathbb{R}^{H \times W} \rightarrow \mathbb{R}^{C_l \times H_l \times W_l}$ denotes a function that resizes the given tensor followed by expansion along the channel dimension of the $l$-th layer. 

Given a pair of feature maps $F_l^q$ and $F_l^s$, we compute a 4D correlation tensor~\cite{min2021hypercorrelation} $\hat{C}_l \in \mathbb{R}^{H_l \times W_l \times H_l \times W_l}$ using cosine similarity:
\begin{equation}
  \hat{C}_l(i,j) = ReLU \left (\frac{F_l^q(i) \cdot \hat{F}_l^s(j)}{\lVert F_l^q(i)\rVert \lVert \hat{F}_l^s(j)\rVert} \right )
\end{equation}
where $i$ and $j$ denote 2D spatial positions of feature maps. We collect correlation tensors computed all the intermediate features of the same spatial size and stack them to obtain a stacked correlation map $\hat{C}_p \in \mathbb{R}^{|\mathcal{L}_p| \times H_p \times W_p \times H_p \times W_p}$, where ($H_p$, $W_p$) are the height and width of the query and support feature maps, and $\mathcal{L}_p$ is a subset of CNN layer indices $\{1,...,L\}$ at pyramid layer $p$, containing correlation maps of identical spatial size.  Given $P$ pyramidal layers, we denote the hypercorrelation pyramid as $\hat{\mathcal{C}} = \{\hat{C}_p\}_{p=1}^P$, representing a collection of feature correlations from multiple visual aspects.

\begin{figure}[h]
  \centering
  \includegraphics[width=\columnwidth]{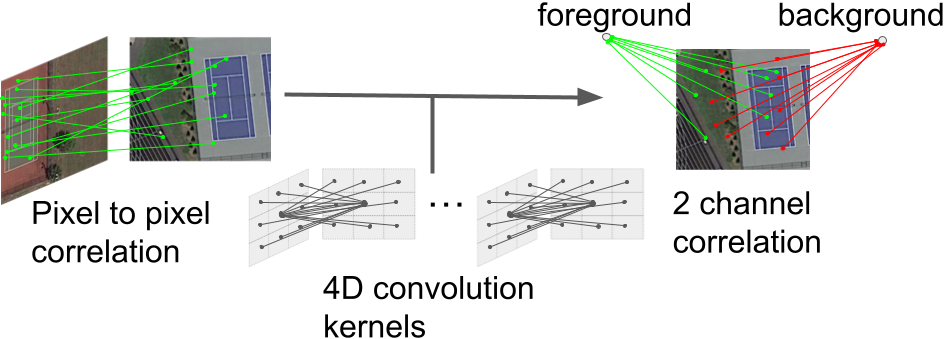}
  \caption{Simplified illustration of the effect of 4D convolution kernels that squeeze the support spatial dimensions.}
  \label{fig:4dconv}
\end{figure}

\noindent \textbf{Correlation Squeeze Encoder.} Our encoder network takes the hypercorrelation pyramid $\hat{\mathcal{C}} = \{\hat{C}_p\}_{p=1}^P$ to effectively squeeze it into a condensed feature map $Z \in \mathbb{R}^{128\times H_1 \times W_1}$. As shown in Figure \ref{fig:4dconv}, sequences of multi-channel 4D convolution with large strides periodically squeeze the last two (support) spatial dimensions of $\hat{C}_p$ down to ($H_\epsilon$, $W_\epsilon$) while the first two spatial (query) dimensions remain the same as ($H_p$, $W_p$). Similar to FPN \cite{lin2017feature} structure, two outputs from adjacent pyramidal layers, $p$ and $p+1$, are merged by element-wise addition after upsampling the (query) spatial dimensions of the upper layer. After merging, the output tensor of the lowest block is further compressed by average-pooling its last two (support) spatial dimensions, which in turn provides a 2-dimensional feature map $Z \in \mathbb{R}^{128 \times H_1 \times W_1}$ that signifies a condensed representation of the hypercorrelation pyramid $\hat{\mathcal{C}}$.

\noindent \textbf{2D-convolutional context decoder.} The decoder network consists of a series of 2D convolutions, ReLU, and upsampling layers followed by sofmax function. The network takes the context representation $Z$ and predicts two-channel map $\hat{M}_{init}^q \in [0, 1]^{2 \times H \times W}$ where two channel values indicate probabilities of foreground and background. Then we take the maximum channel value at each pixel of $\hat{M}_{init}^q$ to obtain initial query mask prediction $\tilde{M}_{init}^q \in \{0, 1\}^{H \times W}$.
 
\subsection{Self-Matching Module}

While the cross-matching module successfully captures intricate correlations between support and query images, it faces limitations when significant disparities exist between the support and query features. Consequently, the initial query mask $\hat{M}_{init}^q$ generated by the cross-matching module may lack crucial details, which is a pain point for the segmentation task. To tackle this issue, 
Self-Matching Module (SMM) is proposed to provide auxiliary support information to segment the query image.

Suppose the query image is $I^q$, and the initial query mask is $\tilde{M}_{init}^q$. In the Self-Matching Module, different from calculating the correlation tensor between masked support features and query features, we calculate the correlation tensor between initial masked query features and query features:

\begin{align}
      \hat{C}_l^{self}(i,j) &= ReLU \left (\frac{F_l^q(i) \cdot \hat{F}_l^q(j)}{\lVert F_l^q(i)\rVert \lVert \hat{F}_l^q(j)\rVert} \right ), \\
  \text{where }
  \hat{F}_q^l &= F_q^l \odot \zeta_l(\tilde{M}_{init}^q)
\end{align}
Following the procedure in Cross-Matching Module, we can obtain $\hat{M}_{self}^q \in [0, 1]^{2 \times H \times W}$. Then, we concatenate $\hat{M}_{self}^q$ with $\hat{M}^q$ and utilize 1x1 conv to reduce the channel dimension to get $\hat{M}_{merge}^q \in [0, 1]^{2 \times H \times W}$. 

In Self-Matching Module, the loss function $\mathcal{L}_{m}$ for training the model can be computed as follows:
\begin{equation}
  \mathcal{L}_m = BCE(\hat{M}_{merge}^q, M^q)
\end{equation}
where $BCE(\cdot)$ is the binary cross entropy loss and $M^q$ is the ground truth mask of the query image.

To further facilitate the Self-Matching procedure, we propose a query self-matching loss:
\begin{equation}
  \mathcal{L}_{aux} = BCE(\hat{M}_{aux}^q, M^q)
\end{equation}
Here, $\hat{M}_{aux}^q$ is generated following the procedure of $\hat{M}_{self}^q$, but with ground truth query mask to calculate the masked query feature $\hat{F}_q^l = F_q^l \odot \zeta_l(M^q)$. The motivation is that the quality of the initial predicted query  mask directly influences the auxiliary information extracted during the self-matching stage. Finally, we train the model in an end-to-end manner by jointly optimizing $\mathcal{L} = \mathcal{L}_m + \lambda \mathcal{L}_{aux}$, where $\lambda$ serves as weight strength, and we set $\lambda = 1.0$ in our experiments. 

\subsection{Spectral Segmentation Module}
Self-Matching Module incorporates the proximity between the initial query mask $\tilde{M}_{init}^q$ and the query image within the model, effectively addressing the challenge of large intra-class variance. However, the performance of this module is influenced by the quality of $\tilde{M}_{init}^q$. To overcome this limitation, we employ a spectral analysis method to extract valuable knowledge from the affinity matrix, which is constructed solely based on the query image. 

\begin{figure}[h]
  \centering
  \includegraphics[width=\columnwidth]{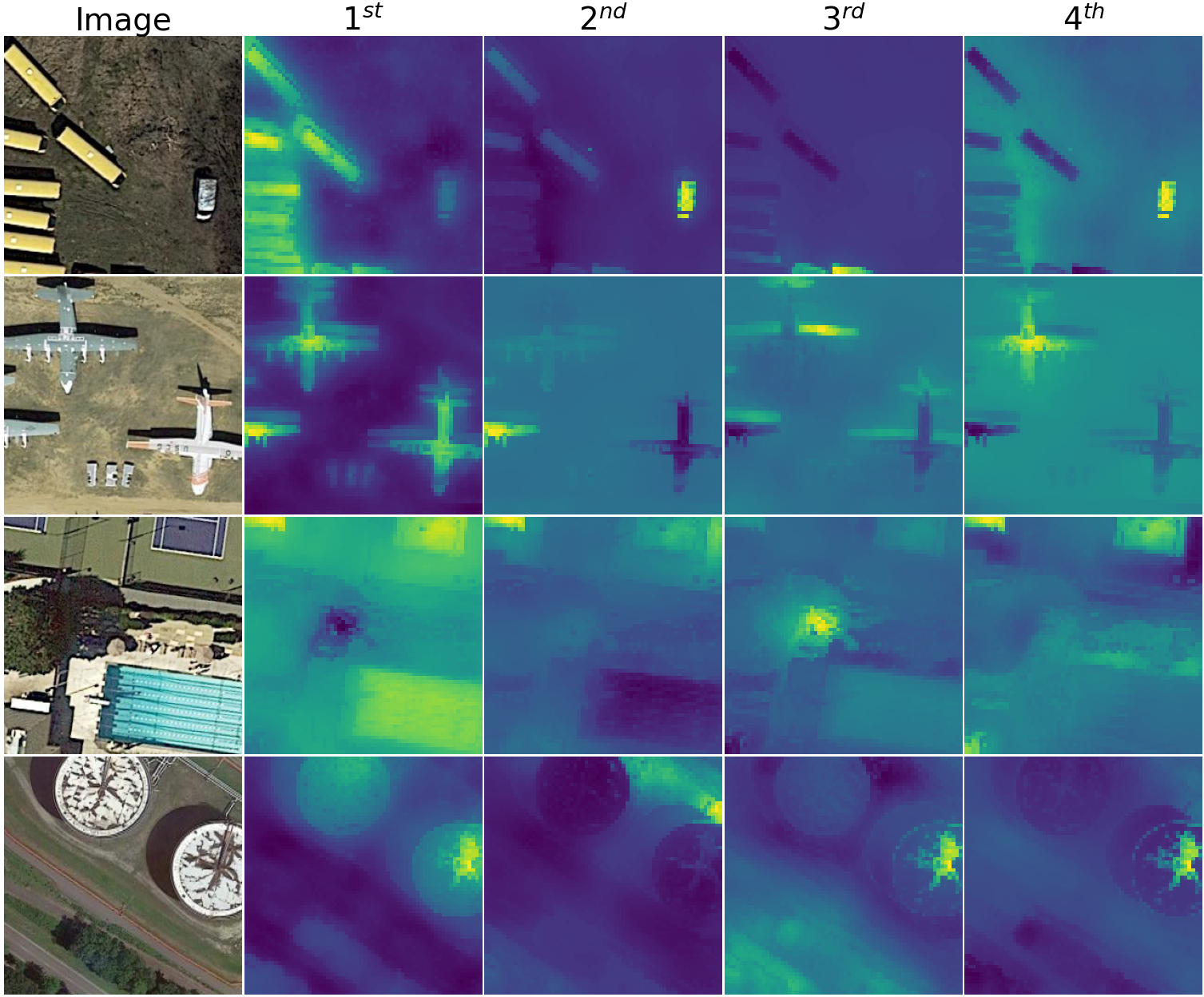}
  \caption{Visualization examples of first four Eigenvectors of our feature
affinity matrix on iSAID-$5^i$ dataset. The
eigenvectors correspond to semantic regions, with the first eigenvector usually identifying the most salient object in the image.}
  \label{fig:eigenvectors}
\end{figure}

The derivation of the affinity matrix is the key to spectral decomposition. Inspired by Melas-Kyriazi et al. \cite{melaskyriazi2022deep}, we leverage the features $f$ from the middle layer of the pretrained backbone to construct an affinity matrix. Additionally, since the features are extracted for aggregating similar features rather than anti-correlated features, we set the affinity thresholding as 0 : 
\begin{equation}
  Z_{sem}(i, j) = f_i f^T_j \odot ( f_i f^T_j > 0)
\end{equation}

While the affinities derived from embedding features are rich in semantic information, it lacks low-level proximity including color similarity and spatial distance. To solve this problem, we adopted image matting \cite{chen2013knn, levin2007closed} to consider the basic visual information in Spectral Segmentation Module. Specifically, we first transform the input image into 
the HSV color space: $X(i)= (cos(h),sin(h),s,v,x,y)_i$, where $h,s,v$ are the respective HSV coordinates and $(x,y)$ denotes the spatial coordinates of pixel $i$. Here $X$ contians color information and position information which can be seen as the 0-th layer feature of the network. Then, we construct a sparse affinity matrix from pixel-wise nearest neighbors based on $X$:
\begin{equation}
  Z_{knn}(i, j) = 
  \begin{cases}
  1 - \lVert X(i) - X(j)\rVert,  & \text{if $i \in KNN_X(j)$} \\
  0 & \text{otherwise}
  \end{cases}
\end{equation}
where $||\cdot||$ denotes 2-norm and $i \in \mathrm{KNN}_X(j)$ are the k-nearest neighbors of j under the distance defined by $X$. The overall affinity matrix is defined as the weighted sum of the two:
\begin{equation}
  Z(i,j) = Z_{sem}(i, j) + \alpha Z_{knn}(i, j)
\end{equation}
The residual ratio $\alpha$ is the hyper-parameter weighing the importance of the visual and semantic information. Empirically, we set $\alpha=5$ in our experiments.

\begin{figure}[h]
  \centering
  \includegraphics[width=\columnwidth]{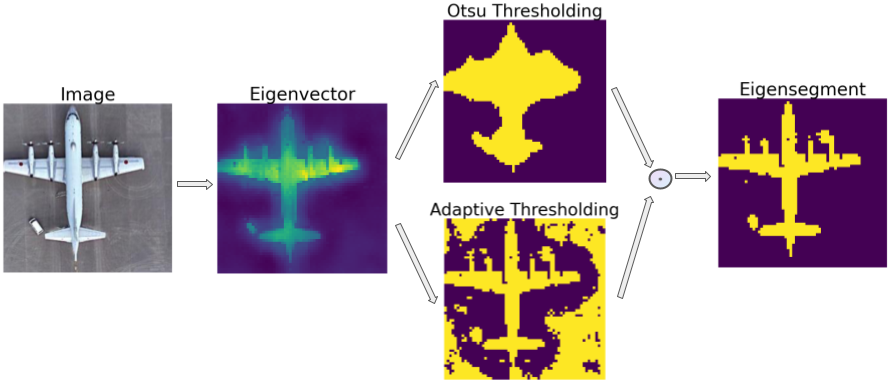}
  \caption{Pipeline of generating eigensegment from an image.}
  \label{fig:eigensegment}
\end{figure}

With the affinity matrix $Z$, we can compute the top $N$ eigenvetors $\{E_i\}_{i=0}^{N-1}$ of the Laplacian $L$. As shown in Figure \ref{fig:eigenvectors}, after being resized to $H \times W$, the eigenvectors are soft segments with continuous values. To convert the soft sements to the hard mask predictions, we propose to introduce two thresholding algorithms into Spectral Segmentation Module. 
The pipeline of this combination process is illustrated in Fig. \ref{fig:eigensegment}.
Specifically, we first utilize Multi-Ostu algorithm~\cite{liao2001fast} to find salient objects and adopt Adaptive Thresholding \cite{gonzalez2009digital} algorithm to 
extract the sharp boundaries in the eignvectors. Then we combine them together with Hadamard product to generate the final eigensegments $\tilde{E}_i \in \{0, 1\}^{H \times W}$:
\begin{equation}
  \tilde{E}_i = MultiOstu(E_i) \odot Adap(E_i), i \in \{1,..,N-1\}
\end{equation}

where we exclude the zero-th constant eigenvector.

\subsection{Inference} \label{fusemodule}
Given a pair of annotated images ${(I^s, M^s), (I^q, M^q)}$, we first generate the predicted query mask $\tilde{M}_{merge}^q$ through Cross-Matching and Self-Matching Modules. Meanwhile, we calculate the top $N - 1$ spectral eigensegments $\{\tilde{E}_i\}_{i=1}^{N-1}$ of each query image. 
Since the eigensegments are class-agnostic, we fuse the merged mask $\tilde{M}_{merge}^q$ with the first eigenvector $\tilde{E}_1$, which has the highest confidence, to obtain the final prediction. 
In addition, to explore the full potential of spectral segmentation, we also present the result of selecting the best $\tilde{E}_{best}$ from $\{\tilde{E}_i\}_{i=1}^{N-1}$ ranked by IoU with ground truth query mask. The final prediction of query mask is a union of $\tilde{E}_{best}$ and $\tilde{M}_{merge}^q$:
\begin{equation}
  \tilde{M}^q = \Phi(\tilde{M}_{merge}^q, \tilde{E}_{best})
\end{equation}
where $\Phi(\cdot)$ is pixel-wise logical \textit{or} function.

Our model can be easily extended to $K$-shot setting: Given $K$ support image-mask pairs $\mathcal{S} = \{(I^s,M_k^s)\}_{k=1}^K$ and a query image $I^q$, the model performs $K$ forward passes to provide a set of $K$ mask predictions $\{\tilde{M}_k^q\}_{k=1}^K$. We perform voting at every pixel location by summing all the $K$ predictions and dividing each output score by the maximum voting score. We assign foreground labels to pixels if their values are larger than some threshold $\tau$ whereas the others are classified as background. We set $\tau = 0.4$ in our experiments.

\section{Experiment}

To demonstrate the effectiveness of the proposed method, the experiments are organized as follows. We first describe the adopted dataset iSAID-$5^i$ and DLRSD-$5^i$. Next, the evaluation metrics and implementation details are introduced. Then, the segmentation results and comparison with the state-of-the-art few-shot segmentation methods are presented. We finally conducted a series of ablation studies to analyze the impact of each component in our proposed method.

\subsection{Datasets}

\begin{table*}[]
  \caption{Classes in iSAID-$5^i$ Dataset}
  \label{tab:isaidclasses}
\resizebox{\textwidth}{!}{\begin{tabular}{ccccccccccccccc}
\hline
C1 &
  C2 &
  C3 &
  C4 &
  C5 &
  C6 &
  C7 &
  C8 &
  C9 &
  C10 &
  C11 &
  C12 &
  C13 &
  C14 &
  C15 \\ \hline
ship &
  \begin{tabular}[c]{@{}l@{}}storage\\ tank\end{tabular} &
  \begin{tabular}[c]{@{}l@{}}baseball\\ diamond\end{tabular} &
  \begin{tabular}[c]{@{}l@{}}tennis\\ court\end{tabular} &
  \begin{tabular}[c]{@{}l@{}}basketball\\ court\end{tabular} &
  \begin{tabular}[c]{@{}l@{}}ground\\ track\\ field\end{tabular} &
  bridge &
  \begin{tabular}[c]{@{}l@{}}large\\ vehicle\end{tabular} &
  \begin{tabular}[c]{@{}l@{}}small\\ vehicle\end{tabular} &
  helicopter &
  \begin{tabular}[c]{@{}l@{}}swimming\\ pool\end{tabular} &
  roundabout &
  \begin{tabular}[c]{@{}l@{}}soccer\\ ball\\ field\end{tabular} &
  plane &
  harbor \\ \hline
\end{tabular}}
\end{table*}

\begin{table*}[]
  \caption{Classes in DLRSD-$5^i$ Dataset}
  \label{tab:dlrsdclasses}
\resizebox{\textwidth}{!}{\begin{tabular}{ccccccccccccccc}
\hline C1       & C2        & C3        & C4   & C5        & C6    & C7   & C8    & C9    & C10         & C11      & C12  & C13 & C14  & C15   \\ \hline
airplane & bare soil & buildings & cars & chaparral & court & dock & field & grass & \begin{tabular}[c]{@{}l@{}}mobile\\ home\end{tabular} & pavement & sand & sea & ship & tanks \\ \hline
\end{tabular}}
\end{table*}

\begin{table}[]
  \caption{Testing Classes for Threefold Cross Validation Test}
  \label{tab:isaidsplits}
\begin{tabular}{ll}
\hline
Dataset & Test classes                                                                                                \\ \hline
iSAID-$5^0$ & \begin{tabular}[c]{@{}l@{}}ship, storage tank, baseball diamond,\\ tennis court, basketball court\end{tabular} \\ 
iSAID-$5^1$ & \begin{tabular}[c]{@{}l@{}}ground track field, bridge, large vehicle,\\ small vehicle, helicopter\end{tabular} \\ 
iSAID-$5^2$ & \begin{tabular}[c]{@{}l@{}}swimming pool, roundabout, soccer ball \\ field, plane, harbor\end{tabular}          \\ \hline
\end{tabular}
\end{table}

\begin{table}[]
  \caption{Testing Classes for Threefold Cross Validation Test}
  \label{tab:dlrsdsplits}
\begin{tabular}{ll}
\hline
Dataset & Test classes                                                                                                \\ \hline
DLRSD-$5^0$ & \begin{tabular}[c]{@{}l@{}}airplane, bare soil, buildings, cars, chaparral\end{tabular} \\ 
DLRSD-$5^1$ & \begin{tabular}[c]{@{}l@{}}court, dock, field, grass, mobile home\end{tabular} \\
DLRSD-$5^2$ & \begin{tabular}[c]{@{}l@{}}pavement, sand, sea, ship, tanks\end{tabular}          \\ \hline
\end{tabular}
\end{table}

\noindent \textbf{iSAID-5$^i$} The iSAID dataset \cite{waqas2019isaid} contains 655,451 object instances for 15 categories across 2,806 high-resolution images, which exactly match the requirement of the few-shot segmentation task. Based on this, Yao et al. \cite{yao2021scale} create the iSAID-$5^i$ dataset following the setting in PASCAL-$5^i$ \cite{shaban2017one}, and the class details are show in Table \ref{tab:isaidclasses}. Particularly, for the 15 object categories in the iSAID-$5^i$ dataset, the cross-validation method is leveraged to evaluate the proposed model by using five classes in one fold as test categories $\mathcal{D}_{test}$ and leveraging the ten classes in the left two folds as the categories of the training set $\mathcal{D}_{train}$. The details of the class splits are shown in Table \ref{tab:isaidsplits}, where $i$ is the fold number. For every fold, we use the same model with the same hyperparameter setup following standard cross-validation protocol. The iSAID-$5^i$ dataset contains 18,076 images for training, 6,363 images for validation and the resolution of all the images is fixed to be $256 \times 256$. Furthermore, this dataset provides sufficient size diversity for the few-shot remote sensing images' semantic segmentation task.

\noindent \textbf{DLRSD-5$^i$} The Dense Labeling Remote Sensing Dataset (DLRSD) \cite{shao2018performance} is a publicly available dataset for evaluating multi-label remote sensing image retrieval and semantic segmentation algorithms. DLRSD contains 2,100 RGB images in total, 17 object classes and the image sizes are fixed as $256 \times 256$ pixel. To balance the number in each fold, we use 15 categories of DLRSD to build DLRSD-5$^i$. The details of the class splits are shown in Table \ref{tab:dlrsdsplits}.

\subsection{Evaluations metrics}
We adopt mean intersection over union (mIoU) as our evaluation metrics. For each category, the IoU is calculated by $\mathrm{IoU} = \frac{TP}{TP + FN + FP}$, where $TP, FN, FP$ respectively denote the number of true positive, false negative and false positive pixels of the predicted mask. The mIoU metric averages over IoU values of all classes in a fold: mIoU = $\frac{1}{C}\sum_{c=1}^C\mathrm{IoU}_c$ where $C$ is the number of classes in the target fold and $\mathrm{IoU}_c$ is the intersection over union of class $c$.

\subsection{Implementation details}

For the backbone network, we employ VGG \cite{simonyan2014very} and ResNet \cite{he2016deep} families pre-trained on ImageNet \cite{deng2009imagenet}, e.g., VGG16, ResNet50, and ResNet101. For VGG16 backone, we extract features after every conv layer in the last two building blocks: from conv4\_x to conv5\_x, and after the last maxpooling layer. For ResNet backbones, we extract features at the end of each bottleneck before ReLU activation: from conv3\_x to conv5\_x. This feature extracting scheme results in 3 pyramidal layers ($P = 3$) for each backbones. In spectral segmentation module, we peek the layer with size $64 \times 64$ as $f$ to construct affinity matrix $Z_{sem}$, which contains rich semantic information and high resolution. The image size in both iSAID-$5^i$ and DLRSD-5$^i$ is $256 \times 256$, i.e., $H,W=256$. This network is implemented in PyTorch \cite{paszke2019pytorch} and optimized with SGD optimizer where the learning rate is 9e-4, the weight decay is 5e-4, and the momentum is 0.9. The learning rate is scheduled with polynomial strategy. The backbone is trained together with 10 times smaller learning rate. 

\subsection{Compared with SOTA}

\begin{table*}[]
\caption{Performance on iSAID-$5^i$ in mIoU. Some results are reported in \cite{yao2021scale}. Numbers in bold indicate the best performance and underlined ones are the second best. Superscript $\dag$ denotes $\tilde{E}_{best}$ is used instead of $\tilde{E}_1$.}
\label{tab:benchmarks}
\begin{tabular}{cc|cccc|cccc|c}
\hline
\multirow{2}{4em}{Backbone network} & \multirow{2}{4em}{Methods} & \multicolumn{4}{c|}{1-shot} &  \multicolumn{4}{c|}{5-shot} & \multirow{2}{4em}{learnable params}                    \\ 
                 &        & fold0 & fold1 & fold2 & mean   & fold0 & fold1 & fold2  & mean   &  \\ \hline
                 & PANet \cite{wang2019panet}  & 17.43 & 11.43 & 15.95 & 14.94      & 17.7  & 14.58 & 20.7   & 17.66      & 14.7M               \\ 
                 & CANet \cite{zhang2019canet}  & 19.73 & 17.98 & 30.93 & 22.88     & 23.45 & 20.53 & 30.12  & 24.70       & 26.4M               \\ 
                 & PMMs \cite{yang2020prototype}  & 20.87 & 16.07 & 24.65 & 20.53      & 23.31 & 16.61 & 27.43  & 22.45    & 25.8M               \\ 
VGG16            & PFENet \cite{tian2020prior} & 16.68 & 15.3  & 27.87 & 19.95     & 18.46 & 18.39 & 28.81  & 21.89  & 10.4M               \\ 
                 & SDM \cite{yao2021scale} & 29.24 & 20.80  & \underline{34.73} & 28.26   & 36.33 & 27.98 & \textbf{42.39}  & \underline{35.57}  & 25.8M               \\
                 & HSNet \cite{min2021hypercorrelation} & 22.74 & 23.05 & 25.76 & 23.84  & 27.20  & 28.86 & 28.82  & 28.29  & 2.6M               \\ \cline{2-11} 
                 & Ours & \underline{30.00}    & \underline{27.41} & 32.43 & \underline{29.94}   & \underline{36.52} & \underline{31.40}  & 37.53  & 35.15 & 5.2M  \\ 
                 & Ours$^{\dag}$ & \textbf{35.71}    & \textbf{30.33} & \textbf{36.68} & \textbf{34.24}   & \textbf{40.40} & \textbf{32.56}  & \underline{39.31}  & \textbf{37.42} & 5.2M  \\ \hline
                 & PANet \cite{wang2019panet}  & 12.36 & 9.11  & 12.05 & 11.17     & 13.82 & 12.4  & 19.12  & 15.11   & 23.5M               \\ 
                 & CANet \cite{zhang2019canet}  & 18.8  & 15.62 & 25.79 & 20.07     & 23.86 & 18.54 & 32     & 24.8   & 20.2M               \\ 
                 & PMMs \cite{yang2020prototype} & 19.02 & 18.51 & 28.42 & 21.98     & 20.89 & 20.87 & 31.23  & 24.33   & 19.6M               \\ 
Resnet50         & PFENet \cite{tian2020prior} & 18.75 & 17.24 & 22.09 & 19.36    & 19.57 & 18.43 & 26.14  & 21.38   & 10.8M               \\ 
                 & SDM \cite{yao2021scale} & 34.29 & 22.25 & 35.62 & 30.72   & 39.88 & \underline{30.59} & 45.70   & 38.72    & 19.6M               \\ 
                 & HSNet \cite{min2021hypercorrelation} & 30.76 & 24.35 & 38.20  & 31.10  & 38.08 & 30.56 & 45.28  & 37.79  & 2.6M                \\ \cline{2-11}
                 & Ours   & \underline{36.21} & \underline{27.42}  & \underline{43.37}  & \underline{35.67}   & \underline{42.58} & 30.30  & \underline{50.26}  & \underline{41.05}  & 5.2M \\
                 & Ours$^{\dag}$   & \textbf{40.74} & \textbf{31.25}  & \textbf{46.40}  & \textbf{39.45}   & \textbf{44.27} & \textbf{31.62} & \textbf{50.45}  & \textbf{42.11}  & 5.2M                \\ \hline
        & HSNet \cite{min2021hypercorrelation} & 34.91 & 26.51 & 40.84 & 34.09  & 41.71 & 31.08 & 48.54  & 40.44  & 2.6M                \\ \cline{2-11}
Resnet101        & Ours   & \underline{37.65} & \underline{29.19} & \underline{42.99} & \underline{36.13} & \underline{41.87} & \underline{32.12} & \textbf{49.63}  & \underline{41.20}  & 5.2M \\ 
                 & Ours$^{\dag}$   & \textbf{40.82} & \textbf{31.38} & \textbf{45.32} & \textbf{39.17} & \textbf{42.52} & \textbf{32.72} & \underline{49.18}  & \textbf{41.47}  & 5.2M                \\ \hline
\end{tabular}
\end{table*}

\begin{table*}[]
\caption{Performance on DLRD-$5^i$ in mIoU. Resnet50 is used as the backbone}
\label{tab:benchmarks2}
\begin{tabular}{c|cccc|cccc}
\hline
\multirow{2}{4em}{Methods} & \multicolumn{4}{c|}{1-shot} & \multicolumn{4}{c}{5-shot} \\
       & fold0 & fold1 & fold2 & mean  & fold0 & fold1 & fold2  & mean  \\ \hline
SDM \cite{yao2021scale}   & 20.11 & 30.84 & 27.87 & 26.27 & 26.03 & 41.74 & 33.55  & 33.77 \\
HSNet \cite{min2021hypercorrelation}  & 22.00 & 47.20 & 34.73 & 34.64 & 27.46 & \underline{52.32} & 46.23  & 42.00 \\ \hline
Ours   & \underline{25.34} & \underline{48.97} & \underline{39.73} & \underline{37.37} & \underline{30.22} & \textbf{52.40} & \underline{47.15}  & \textbf{43.26} \\
Ours$^{\dag}$  & \textbf{26.48} & \textbf{49.59} & \textbf{41.89} & \textbf{39.32} & \textbf{30.26} & 51.08 & \textbf{47.60}  & \underline{42.92} \\ \hline
\end{tabular}
\end{table*}

To assess the efficacy of our model, we extensively compare it with state-of-the-art (SOTA) methods \cite{yang2020prototype, tian2020prior, zhang2019canet, wang2019panet, min2021hypercorrelation, yao2021scale} on the iSAID-$5^i$ and DLRSD-5$^i$ dataset, employing different backbone networks and few-shot settings. 

\noindent \textbf{iSAID-5$^i$} Table \ref{tab:benchmarks} presents a summary of the results on iSAID-5$^i$. When using $\tilde{E}_1$, our method outperforms other state-of-the-art methods in almost all the experiment settings. Notable, with Resnet50 as backbone, our method achieves 4.57\% and 2.33\% improvement in mIoU over the state-of-the-art in the 1-shot setting and 5-shot setting respectively. When $\tilde{E}_{best}$ is used, the improvement is further enlarged and comes to 8.35\% and 3.39\%.

\noindent \textbf{DLRSD-5$^i$} Table \ref{tab:benchmarks2} presents a summary of the results on DLRSD-5$^i$. Resnet50 is used as the backbone. When $\tilde{E}_1$ is used, our method achieves 2.73\% and 1.26\% improvement in 1-shot setting and 5-shot setting respectively. When $\tilde{E}_{best}$ rather than $\tilde{E}_1$ is used, out method achieves 4.68\% improvement over the state-of-the-art in the 1-shot setting.

To conduct a more thorough analysis of the performance across diverse classes in the few-shot setting, we have gathered detailed results for the one-shot scenario, utilizing the ResNet50 \cite{he2016deep} backbone. The specific outcomes are presented in Table \ref{tab:perclasses} and \ref{tab:dlrsdperclasses} on iSAID-5$^i$ and DLRSD-5$^i$. On both datasets, our model demonstrates the highest performance when compared to other state-of-the-art (SOTA) methods in 10 out of 15 categories, while in the remaining classes, our model achieves the second-best performance. This substantiates the effectiveness and versatility of our approach.

Notably, we observe an intriguing trend where the improvement in the 1-shot setting is more significant than that in the 5-shot setting across all three backbones. This observation aligns with our design choice, suggesting that our method effectively mitigates intra-class variation. Conversely, in the 5-shot setting, it is more likely that some support images closely resemble the query image. 

\begin{figure*}[h]
  \centering
  \includegraphics[width=\linewidth]{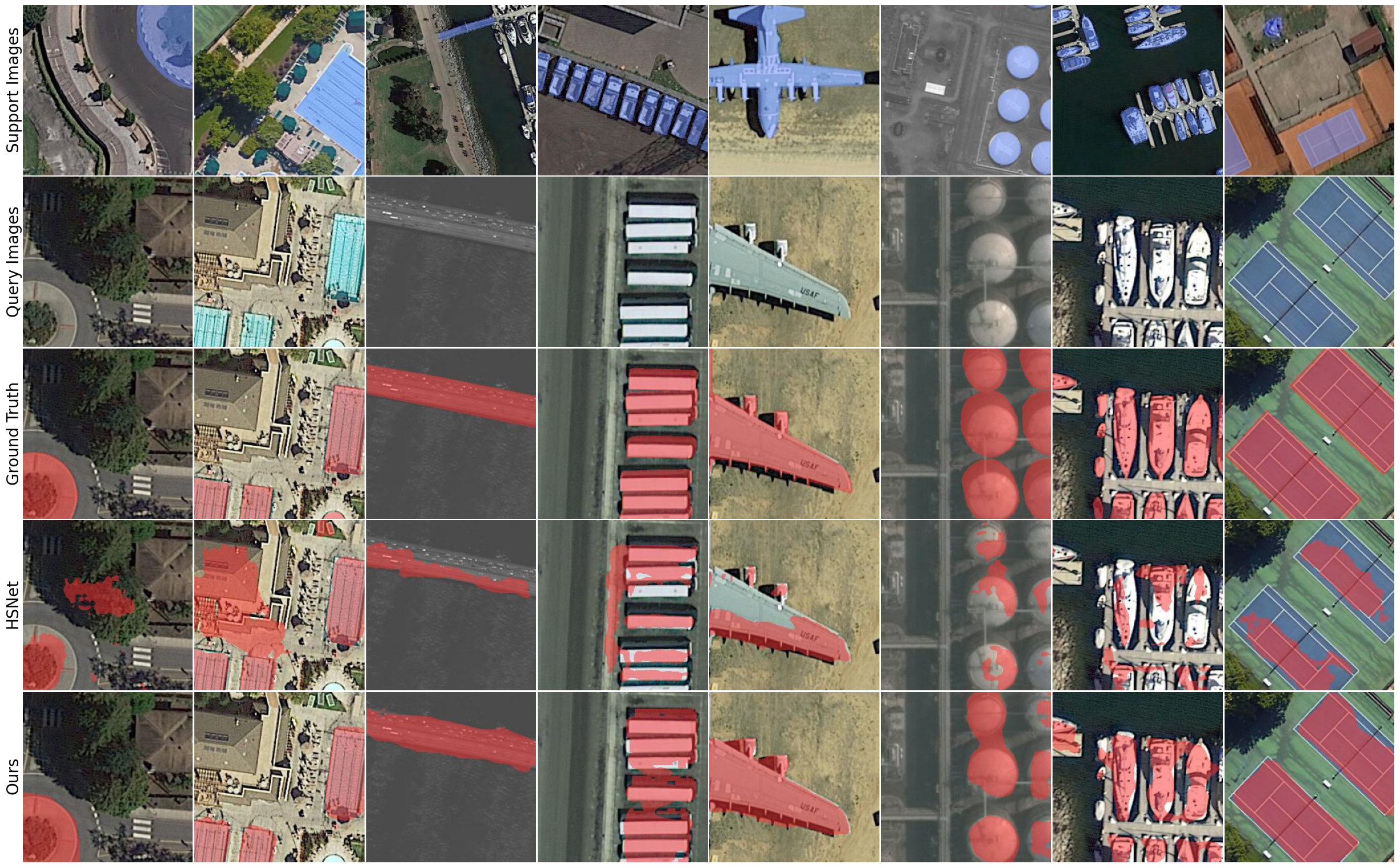}
  \caption{Qualitative results of 1-shot settings on iSAID-$5^i$ dataset.}
  \label{fig:qualitative}
\end{figure*}

Considering the extensive analysis conducted, we can confidently conclude that our proposed method effectively tackles the few-shot semantic segmentation task for remote sensing images. Qualitative results are shown in Fig \ref{fig:qualitative}.

\begin{table*}[]
\caption{Performance comparisons of diverse classes on the iSAID-$5^i$ dataset with 1-shot setting}
\label{tab:perclasses}
\begin{tabular}{c|ccccccccccccccc}
\hline
Methods &
  C1 &
  C2 &
  C3 &
  C4 &
  C5 &
  C6 &
  C7 &
  C8 &
  C9 &
  C10 &
  C11 &
  C12 &
  C13 &
  C14 &
  C15 \\ \hline
SDM \cite{yao2021scale} &
  \textbf{37.66} &
  \underline{34.37} &
  34.45 &
  39.81 &
  \textbf{25.14} &
  16.77 &
  34.53 &
  \underline{30.50} &
  \underline{12.42} &
  17.02 &
  20.69 &
  \textbf{56.83} &
  42.80 &
  \underline{40.52} &
  \underline{17.26} \\
HSNet \cite{min2021hypercorrelation} &
  18.93 &
  30.01 &
  \underline{37.60} &
  \underline{45.33} &
  21.95 &
  \underline{25.11} &
  \textbf{37.17} &
  27.43 &
  11.03 &
  \underline{21.01} &
  \underline{32.22} &
  50.07 &
  \textbf{54.27} &
  37.98 &
  16.46 \\ \hline
Ours &
  \underline{26.76} &
  \textbf{43.42} &
  \textbf{40.27} &
  \textbf{46.74} &
  \underline{22.10} &
  \textbf{27.37} &
  \underline{36.75} &
  \textbf{32.94} &
  \textbf{14.53} &
  \textbf{23.89} &
  \textbf{46.85} &
  \underline{55.06} &
  \underline{45.77} &
  \textbf{48.02} &
  \textbf{23.30} \\ \hline
\end{tabular}
\end{table*}

\begin{table*}[]
\caption{Performance comparisons of diverse classes on the DLRSD-$5^i$ dataset with 1-shot setting}
\begin{tabular}{c|ccccccccccccccc}
\hline Methods & C1    & C2    & C3    & C4    & C5    & C6    & C7    & C8    & C9    & C10   & C11   & C12   & C13   & C14   & C15   \\ \hline
SDM \cite{yao2021scale}    & 5.51  & \underline{22.74} & \textbf{29.00} & 3.83  & \underline{39.49} & 5.30  & 19.97 & 84.13 & 8.94  & \underline{35.90} & 11.96 & \textbf{31.99} & 49.03 & \underline{38.79} & 7.57  \\
HSNet \cite{min2021hypercorrelation}  & \textbf{23.63} & 18.39 & 21.41 & \underline{8.55} & 38.02 & \underline{63.45} & \underline{24.56} & \textbf{96.49} & \underline{18.33} & 33.20 & \underline{20.88} & 24.66 & \underline{57.13} & 35.00 & \underline{35.94} \\ \hline
Ours    & \underline{23.58} & \textbf{25.32} & \underline{26.99} & \textbf{10.45} & \textbf{40.37} & \underline{53.27} & \textbf{25.49} & \underline{96.29} & \textbf{29.10} & \textbf{40.68} & \textbf{30.09} & \underline{24.80} & \textbf{60.07} & \textbf{46.72} & \textbf{37.00} \\ \hline
\end{tabular}
\label{tab:dlrsdperclasses}

\end{table*}

\subsection{Ablation study}

\noindent \textbf{Ablation study on designed modules.} To further demonstrate the effectiveness of our designed modules, we conduct ablation experiments on iSAID-5$^i$ using the 1-shot setting and ResNet50 backbone. Table \ref{tab:components} presents the results obtained. The baseline model solely comprises the Cross-Matching Module, which is based on HSNet \cite{min2021hypercorrelation}. By introducing the Self-Matching Module, we observe a notable improvement of 3.95\% in mIoU. This outcome highlights the significant benefit derived from the Self-Matching Module, which introduce proximity information between initial query mask and query image into the model.

\begin{table}[]
\caption{Ablation study on the designed components of the proposed algorithm}
\label{tab:components}
\begin{tabular}{ccc|cccl}
\hline
Self-Matching &
  $\tilde{E}_1$ &
  $\tilde{E}_{best}$  &
  fold0 &
  fold1 &
  fold2 &
  mIoU  \\ \hline
 $\times$ & $\times$  &  $\times$ & 30.76 & 24.35 & 38.20  & 31.10  \\ \hline
  $\times$ & $\checkmark$  &  $\times$ & 32.65 & 25.34 & 39.12 & $32.37_{\uparrow1.27}$  \\ \hline
\checkmark & $\times$  &  $\times$ & 34.64 & 26.85 & 43.36 & $35.05_{\uparrow3.95}$ \\ \hline
\checkmark & \checkmark &  $\times$ & 36.21 & 27.42 & 43.37 & $35.67_{\uparrow4.57}$  \\ \hline
\checkmark & $\times$  & \checkmark & 39.80  & 29.70  & 48.19 & $39.23_{\uparrow8.13}$ \\ \hline
\end{tabular}
\end{table}

\noindent \textbf{Ablation study on fusion strategy of eigensegments.} As shown in Table \ref{tab:components}, when we fuse $\tilde{E}_1$ with $\tilde{M}_{init}^q$ generated by the Cross-Matching Module, we achieve a notable improvement of 1.27\% in mIoU, which proves the efficacy of the Spectral Segmentation Module. When we fuse $\tilde{E}_1$ with $\tilde{M}_{merge}^q$ generated by Self-Matching Module, the total improvement comes to 4.57\%, which is a large margin. In addition, our investigation reveals that the target object is not always contained within the first eigensegment, as it may not be the most salient foreground object. For instance, in the first image of Figure \ref{fig:eigenvectors}, the buses are the most salient objects and they are present in the first eigenvector, while the small vehicle is present in the second eigenvector. To fully explore the capabilities of spectral segmentations, as discussed in Section \ref{fusemodule}, we fuse $\tilde{E}_{best}$ with $\tilde{M}_{merge}^q$. This operation yields a significant increase in improvement, with a difference of 8.13\% from the baseline. This result demonstrates that the Spectral Segmentation Module, which solely mine knowledge from the query image, successfully tackles the large discrepancies between the support and query image observed in remote sensing images.

\begin{table}[]
\caption{Ablation study on design of Self-Matching Module. In single-branch design, we use same 4D conv kernels in both Cross-Matching and Self-Matching Module.}
\label{tab:shared}
\begin{tabular}{c|cccc}
\hline
Experiments & fold0 & fold1 & fold2 & mIoU \\ \hline
HSNet \cite{min2021hypercorrelation}   & \underline{30.76} & \underline{24.35} & 38.20  & \underline{31.10}  \\ \hline
single-branch   & 26.12 & 25.77 & \underline{38.82} & 30.24 \\ \hline
two-branch & \textbf{34.64} & \textbf{26.85}  & \textbf{43.36}  & \textbf{35.05}  \\ \hline
\end{tabular}
\end{table}

\noindent \textbf{Ablation study on design of Self-Matching Module.} In our model architecture, we employ a two-branch network, where the Cross-Matching Module and Self-Matching Module have separate weights. This choice doubles the number of learnable parameters in our model. To investigate the possibility of reducing memory consumption, we conduct an ablation study on a single-branch structure, where the Cross-Matching Module and Self-Matching Module share the same weights. However, as shown in Table \ref{tab:shared}, the performance of the single-branch structure is even inferior to that of HSNet \cite{min2021hypercorrelation}, not to mention the two-branch network. This observation suggests that the Cross-Matching Module and Self-Matching Module have subtle differences, and sharing weights actually harms the performance of the Cross-Matching Module instead of enhancing it. Nevertheless, due to the sparse design of center-pivot 4D convolution \cite{min2021hypercorrelation} we adopt, our model still has a relatively small number of learnable parameters compared to other methods \cite{yao2021scale, yang2020prototype, tian2020prior, zhang2019canet, wang2019panet}. 

\begin{table}[]
\caption{Ablation study on the hyperparameter $\alpha$ in the Spectral Segmentation Module.}
\label{tab:ablamba2}
\begin{tabular}{c|cccl}
\hline
$\alpha$ & fold0 & fold1 & fold2 & mIoU  \\ \hline
1       & 36.62 & 27.50 & 42.63 & $35.58_{\uparrow0.63}$ \\ 
5       & 36.21 & 27.42 & 43.37 & \textbf{$35.67_{\uparrow0.72}$} \\
10      & 35.91 & 27.41 & 43.41 & $35.58_{\uparrow0.63}$ \\
20      & 35.86 & 27.10 & 43.80 & $35.58_{\uparrow0.63}$ \\
50      & 35.65 & 26.96 & 43.42 & $35.34_{\uparrow0.39}$ \\ \hline
\end{tabular}
\end{table}

\noindent \textbf{Ablation study on $\alpha$ of spectral segmentation module.} In the spectral segmentation module, $\alpha$ is a key hyperparameter to balance the semantic affinity matrix $Z_{sem}$ and $Z_{knn}$ which contains raw image information. To select the best value of $\alpha$, we construct some ablation studies on iSAID-5$^i$ with 1-shot setting and Resnet50 backbone. As shown in Table \ref{tab:ablamba2}, $\alpha = 5$ achieves the best performance.

\section{Conclusion}
In this work, we propose a novel \emph{SCCNet} for the few-shot remote sensing image semantic segmentation task. 
Specifically, Self-Matching Module is designed to incorporate the initial query mask prediction to collect query features in high-confidence regions and then use the generated query prototype to perform self-matching with query features.
In addition, we propose the Spectral Segmentation Module with spectral analysis methods to produce class-agnostic segmentations of query images without the supervision of any annotations.
The proposed model is evaluated on two commonly adopted benmarks for few-shot remote sensing image semantic segmentation.
Without any extra knowledge or data information, our \emph{SCCNet} outperforms previous work by a large margin.

\bibliographystyle{ACM-Reference-Format}
\bibliography{reference}






\end{document}